\documentclass[runningheads,a4paper]{llncs}

\usepackage{times}
\usepackage{amssymb}
\usepackage{graphicx}
\usepackage{url}
\usepackage{hyperref}

\newcommand{\keywords}[1]{\par\addvspace\baselineskip
\noindent\keywordname\enspace\ignorespaces#1}

\urldef{\mailsa}\path|{keara.schaaij, iris.hendrickx}@ru.nl|
\urldef{\mailsb}\path|{roel.boumans, tibor.bosse}@ru.nl|
\begin{document}

% TSD 2025: Put your title here (please, use capitalization, see e.g.
% webiste TSD: https://www.kiv.zcu.cz/tsd2025/ 
\title{Towards Stable and Personalised Profiles for Lexical Alignment in Spoken Human-Agent Dialogue%
\thanks{
This preprint has not undergone peer review or any post-submission improvements or corrections. The Version of Record of this contribution is published in TSD 2025. Lecture Notes in Computer Science, vol 16029 and is available online at \href{https://doi.org/10.1007/978-3-032-02548-7_5}{https://doi.org/10.1007/978-3-032-02548-7\_5}}}

% TSD 2025: a short form should be given in case the title is too long for the running head
\titlerunning{Towards Stable and Personalised Profiles for Lexical Alignment}
%\titlerunning{ Personalised Lexical Profiles for Lexical Alignment}
% TSD 2025: Author's names. Chinese authors should write their first names(s)
% in front of their surnames. This ensures that the names appear correctly in
% If the names contain accented characters, please use escape codes
% (refer to http://en.wikibooks.org/wiki/LaTeX/Special_Characters#Escaped_codes)
% \author{Firstname1 Surname1 \and Firstname2 Surname2 \and Firstname3 Surname3}
% \author{Keara Schaaij and Roel Boumans and Tibor Bosse and Iris Hendrickx}

% TSD 2025: For authors from different institutions, please use the following
% form including institution reference
\author{Keara Schaaij\inst{1}\orcidID{0009-0002-8385-1654} \and 
Roel Boumans\inst{2}\orcidID{0000-0002-1287-6363} \and
Tibor Bosse\inst{2}\orcidID{0000-0003-4233-0406}\and
Iris Hendrickx\inst{1}\orcidID{0000-0002-9351-6449}}

% TSD 2025: Author's names for headings. For 1-2 authors, use the following form
% \authorrunning{Firstname1 Surname1 and Firstname2 Surname2}
% TSD 2025: For more than 2 authors, please, use the following
%\authorrunning{Name1 Surname1 et al.}
\authorrunning{Keara Schaaij et al.}

% % TSD 2025: The authors' affiliations
% \institute{Affiliation1, Institute1, Address \\
% % TSD 2025: optional url
% \url{www.website.org} \\
% \mailsa\\
% % TSD 2025: For authors from different institutions, add 2nd institution, etc.
% % \and
% % Affiliation2, Institute2, Address \\
% % \url{www.website.org} \\
% % \mailsb\\
% }

\institute{Centre for Language Studies, Centre for Language and Speech Technology, Radboud University,Nijmegen, The Netherlands \\
\mailsa\\
\and
Behavioural Science Institute, Radboud University, Nijmegen, The Netherlands \\
\mailsb\\}

% TSD 2025: Put all authors' names to the proceeding index (surname, first name)
\index{Schaaij, Keara}
\index{Boumans, Roel}
\index{Bosse, Tibor}
\index{Hendrickx, Iris}

\toctitle{} \tocauthor{}

\maketitle

%
%
%	TSD 2025 SUBMISSION TEXT
%
%
\begin{abstract}
% TSD 2025:
Lexical alignment, where speakers start to use similar words across conversation, is known to contribute to successful communication. However, its implementation in conversational agents remains underexplored, particularly considering the recent advancements in large language models (LLMs). Following strategies used in personalising conversational agents, this study investigates the construction of stable, personalised lexical profiles as a foundation for lexical alignment in human-agent dialogue. Specifically, we varied the amounts of transcribed spoken data used for construction as well as the number of items included in the profiles per part-of-speech (POS) category and evaluated profile performance across time using recall, coverage, and cosine similarity metrics. It was shown that smaller and more compact profiles, created after 10 minutes of transcribed speech containing 5 items for adjectives, 5 items for conjunctions, and 10 items for adverbs, nouns, pronouns, and verbs each, offered the best balance in both performance and data efficiency. In conclusion, this study offers practical insights to construct stable, personalised lexical profiles, taking into account minimal data requirements and offering an important step toward lexical alignment strategies in conversational agents.

% TSD 2025: keywords, comma-separated
\end{abstract}

\section{Introduction}
Alignment is the concept in which two speakers start to use similar linguistic representations during dialogue \cite{pickering2004toward}. It has been argued to be essential for successful communication and can occur at three levels: lexical (word choice), syntactic (sentence structure), and semantic (meaning) \cite{pickering2004toward}. Given the importance of alignment in human-human dialogue, there has been an increased interest in investigating this phenomenon in human-agent dialogue. Research has confirmed its presence in human-agent interactions \cite{sinclair2019tutorbot,koulouri}, and various methods have been proposed to let agents align lexically with users \cite{nunez2023virtual,thomas2018style}. Overall, lexical alignment in human-agent dialogue has been found to positively impact the interaction with and perception of the agent \cite{nunez2023virtual,thomas2018style}.

With the advancements in the development and use of large language models (LLMs), there has been an interest in leveraging these models to generate responses aligning with specific linguistic styles \cite{chen2024persona,favela2023}. For example, LLMs can be prompted to respond in a way a medical professional or well-known figure such as Albert Einstein would respond \cite{chen2024persona}. Moreover, incorporating individual-specific information into the prompts, such as personality type or reading grade level, has been shown to influence the language used in the responses generated by the models \cite{amin2024assessing,chen2024persona,jiang2023personallm}. A few studies have explored prompting strategies to achieve lexical alignment between the user input and LLM-generated responses \cite{srivastava2024}. While these approaches are promising, they often relied on the informativeness of user input during the interaction, which poses a limitation in contexts where user input may be limited, inconsistent, or cognitively demanding for the user. 

Communication with people with dementia (PwD) is an example of such a case where relying on real-time language input is impractical. Specifically, PwD often experience difficulties in communication, such as word-finding problems or unintentionally substituting words \cite{alzheimers-society-2024,banovic2018communication}. The present study is part of a broader initiative aimed at developing and evaluating a voice-based AI system for Dutch elderly people with dementia (PwD), with the goal of alleviating the burden on carers. Specifically, the research focuses on adapting a virtual human framework to enable meaningful and positive interactions with PwD, with the implementation of lexical alignment as a first step \cite{boumans2024virtual,boumans2024voice}. This use case underscores the relevance of exploring lexical alignment techniques that do not depend on extensive user input at times of the interaction. 

Therefore, the current study addresses the question: \textit{How can we construct stable, personalised lexical profiles that form the basis for lexical alignment in human-agent dialogue?} Specifically, two primary objectives are explored: first, to determine the amount of conversational data needed to construct lexical profiles that accurately represent an individual's speech patterns and second, to assess the temporal stability of these profiles. To achieve this, lexical profiles were created using varying amounts of conversational data from individual speakers and by varying the size of these profiles.

\section{Related work}\label{Related work}
\subsection{Alignment in Human-Human Dialogue}\label{alignment in h-h dialogue}

It is well established that two speakers tend to naturally align their ways of speaking during conversations by starting to use similar linguistic patterns \cite{pickering2004toward}. The Interactive Alignment Model proposed by Pickering and Garrod \cite{pickering2004toward} suggests that alignment occurs at multiple linguistic levels, including the lexical (word choices), syntactic (sentence structure), and semantic levels (meaning).

Lexical alignment refers to the tendency of speakers to use similar terms and phrases for underlying concepts \cite{pickering2004toward,sinclair2019tutorbot,srivastava-2024-overview}. For example, if one person refers to an object as a \textit{chair}, their conversational partner will also refer to the object as a \textit{chair}, even if they would normally refer to it as a \textit{seat} \cite{sinclair2019tutorbot}. Syntactic alignment occurs when two speakers match on sentence structure: if one speaker says, \textit{"I want to have a movie night this evening"}, then the conversational partner would respond, \textit{"I am excited to have a movie night this evening"}, rather than restructuring the sentence to \textit{"Having a movie night this evening seems exciting"} \cite{sinclair2019tutorbot}. Semantic alignment refers to the similar use of higher levels of representations between two speakers, such as dialogue acts \cite{sinclair2019tutorbot}.

Research has shown that lexical alignment is influenced by how speakers perceive their conversational partner, particularly their perceived linguistic proficiency. Zhenguang et al. \cite{zhenguang} observed increased lexical alignment during a picture-naming task when the partner was a non-native speaker or child. Similarly, Suffil et al. \cite{suffill2021lexical} found increased alignment during a map-based navigation task in case their partner was a non-native speaker. Ivanova et al. \cite{ivanova2021lexical} also found increased alignment in a picture-matching game when native speakers interacted with non-native speakers, especially in cases where communicative success was not guaranteed. Together, these findings suggest speakers increase lexical alignment to support effective communication, especially in case they perceive their partner as less linguistically competent. This perception is often a result of the reduced fluency and clarity in their partner's speech, generally stemming from word-finding difficulties \cite{ivanova2021lexical}. 

This is particularly relevant in the context of dementia speech, as it closely resembles the challenges in communication experienced by PwD \cite{alzheimers-society-2024,banovic2018communication}. Therefore, lexically aligning with PwD may not only support effective communication but also reflect the natural adjustments that occur during dialogue. In addition, research on dementia speech classification has identified various linguistic features that differentiate dementia speech from healthy speech. Specifically, dementia speech was shown to be characterised by a decline in noun usage and an increased use of pronouns and adjectives \cite{bucks-2000,williams-2023}. In addition, it was found PwD tend to use more conjunctions, potentially reflecting hesitation in speech production \cite{hernandezdominguez-2018}. Verb usage was also found to change, both in terms of the types of verbs used and their frequency \cite{bucks-2000,rodriguez-2024}. These differences further underscore the relevance of lexical alignment with PwD during conversation.

\subsection{Lexical Alignment in Human-Agent Dialogue}
In recent years, there has been a growing interest in implementing lexical alignment in human-agent dialogue based on its importance for successful communication. Prior work, such as that by Sinclair et al. \cite{sinclair2019tutorbot} and Koulouri et al. \cite{koulouri}, has explored the presence of lexical alignment in human-agent dialogue. Additionally, several studies have shown positive influences of lexical alignment on the interaction between human and agent as well as on the perception of the agent \cite{levitan2016implementing,nunez2023virtual}.

Existing implementations of lexical alignment in conversational agents have predominantly relied on partial rule-based and intent-based approaches \cite{duplessis,srivastava-2024-overview}. Specifically, response generation often combined predefined templates and rules with dynamic adjustments based on the inferred goal of the user and incorporated relevant lexical information from the user input into the templates \cite{duplessis,srivastava-2024-overview}. However, the recent advancements in the use and development of LLMs have expanded the capabilities of conversational agents by increasing the flexibility and generalisability of the generated responses, moving away from rule-based and intent-based methods \cite{favela2023}. 

Srivastava et al. \cite{srivastava2024} integrated lexical alignment directly into the response generation of an LLM (GPT-4 turbo) in a travel planning task. In their study, the agent's responses were automatically generated by the model, corrected using a Wizard of Oz (WOZ) approach, after which the model was used again to either lexically align or misalign the responses with the user's input. To achieve lexical alignment, the LLM was prompted to maximise the match with the linguistic style of the user by using similar words and idioms as in the provided user input and to minimise the match to achieve lexical misalignment. Lexical alignment in dialogue was calculated using a subset of metrics from the method of Duplessis et al. \cite{duplessis}. Significant differences in alignment scores across the two conditions were observed, confirming the effectiveness of (mis)alignment with the users in the generated responses. 

\subsection{Personalisation of LLMs}
Despite the advancements in using LLMs in conversational agents, research explicitly focusing on lexical alignment in response generation remains limited. However, various other personalisation techniques for LLMs have gained attention. Particular relevant is the concept of Role Playing Language Agents (RPLAs), which enable LLMs to replicate specific personas. Chen et al. \cite{chen2024persona} categorise three types of personas. The first is the Demographic Persona, in which prompts are used to let the LLM mimic the speaking style of a stereotypical group of people, such as a teacher, leveraging inherent stereotypical persona knowledge in the LLM \cite{chen2024persona}. Second is the Character Persona, where LLMs are prompted to speak or act as well-known figures, like Harry Potter, characterised by definitive attributes and narratives \cite{chen2024persona}. Third, and most relevant to lexical alignment, is the Individualised Persona, in which a personal profile is built and being updated during interactions, emphasising the unique experiences, needs, and preferences of each individual \cite{chen2024persona}. 

Following the notion of personalising LLM-generated responses using prompt engineering, Jiang et al. \cite{jiang2023personallm} explored how the models GPT-3.5 and GPT-4 reflect various personality traits in their responses by incorporating a persona description in the prompt. Their results demonstrated the ability of these LLMs to express specified personality types in a personality assessment, as well as in linguistic style. More closely related to simulating specific language use is the study by Reichenpfader et al. \cite{reichenpfader2024large}, where an LLM (GPT-4) simulated patient questions corresponding to different levels of health literacy (related to how individuals obtain, use, and understand health-related information). Specifically, the LLM was prompted to provide alternative questions based on an example question and a patient vignette describing their health literacy level. Similarly, Amin et al. \cite{amin2024assessing} examined whether LLMs could generate responses tailored to specific reading grade levels as indicated in the prompt. Their results showed that LLMs could adjust their output based on the reading grade level to a certain extent. However, accurately targeting the specific level, particularly below the 6th grade level, proved difficult.

In summary, prior work has demonstrated the feasibility of personalisation in the response generation by conversational agents relying on LLMs. The personalised profiles used by RPLA's Individualised Personas present a promising framework to achieve lexical alignment by constructing profiles that incorporate the lexical characteristics of individual users. However, to the best of our knowledge, research on the use and creation of such profiles specifically for lexical alignment remains limited. Therefore, there is a need to explore how we can construct such profiles and evaluate their stability and relevance over time as a basis for lexical alignment in human-agent dialogue. Such efforts will provide insights into the efficient construction of stable, personalised lexical profiles. 

\section{Method}
To address the first objective of this study, determining the amount of data needed to construct lexical profiles, we ran experiments using varying amounts of transcribed spoken data and explored different profile sizes by adjusting the number of included items to examine which configuration for the profiles represented the speakers best. For the second objective, assessing the temporal stability of these profiles, we evaluated the profiles on future spoken data from the same person by applying commonly used repetition-based metrics in lexical alignment research to assess their stability over time. 

\subsection{Data}
Since the overarching research focuses on voice-based interactions with Dutch elderly PwD, the dataset was selected taking this context into account. At the time of the study, no dataset on dementia speech for Dutch was available. As a result, spontaneous speech from elderly individuals was chosen, reflecting natural speech patterns and age-relevant speech characteristics. To this end, the \textit{Interview Collectie Nederlandse Veteranen} (ICNV), provided by the Netherlands Veterans Institute, was used \cite{nederlands-veteraneninstituut-2024}. This collection contains interviews with Dutch veterans who have participated in wars, armed conflicts, or peace missions involving the Netherlands since 1940, in which they share their personal experiences.

Specifically, the transcripts of 50 interviews involving 38 unique interviewees (36 male and 2 female) were used. The speakers' ages ranged from 75 to 95, with an average of 85 years old; however, age information was unclear for two speakers. A subset of 10 randomly chosen interview transcripts was used as the test set. The average duration of the used interviews was 124 minutes, ranging from 50 minutes to 272 minutes. Due to privacy restrictions, only the transcripts were used, not the corresponding audio files. The transcripts included time markers at five-minute intervals and included common features of spontaneous speech, such as long pauses (marked by ellipses, ``...") and interrupted words or sentences (marked by hyphens, ``-"). To facilitate accurate tokenisation and sentence segmentation, a preprocessing step was applied in which ellipses were replaced with the marker ``PAUSE" and hyphens with ``BREAK". Given the goal of constructing personalised lexical profiles, only the transcribed speech of the interviewee was analysed, and that of the interviewer was excluded.

\subsection{Lexical Features}
The lexical profile integrated lexical features to represent the speaker's preferred vocabulary and recurring lexical patterns, reflecting the components of lexical alignment. Specifically, it included the most frequently used terms per part-of-speech (POS) category and the most common word-based n-grams, capturing the core language characteristics of the speaker's language. The selected POS categories included nouns, pronouns, adjectives, conjunctions, verbs, and adverbs, which largely reflect the categories affected in dementia speech discussed in Section \ref{alignment in h-h dialogue}. 

For the vocabulary items, the five most frequently used terms per POS category were included as a baseline; however, additional configurations using values of three, 10, 15, and 20 were also explored to assess their influence on performance. To ensure the relevance of the incorporated terms and prevent the inclusion of sparse examples, a minimum occurrence threshold of five was applied. To extract the vocabulary of the interviewee, SpaCy's \textit{nl\_core\_news\_lg} pipeline for the Dutch language was used for POS tagging and tokenisation \cite{Spacy}.

In addition to common vocabulary items per POS category, the word-based n-grams were extracted, a well-known method to identify recurring patterns in dialogue \cite{cappelle2016towards}. A maximum of three n-grams were extracted for values of \textit{n} ranging from two to five, consistent with ranges in Dutch n-grams datasets, such as Dutch Twitter data \cite{bouma2015n} and the Web 1T 5-gram corpus created by Google \cite{brants2009web}. An inclusion threshold of three was applied to prevent the inclusion of sparse examples. It is important to note that the n-grams extracted retained annotations for pauses ("PAUSE") and interruptions ("BREAK"), which were manually addressed during preprocessing.

\subsection{Timeframes}
\begin{figure}[ht]
\centering
\includegraphics[width=\linewidth]{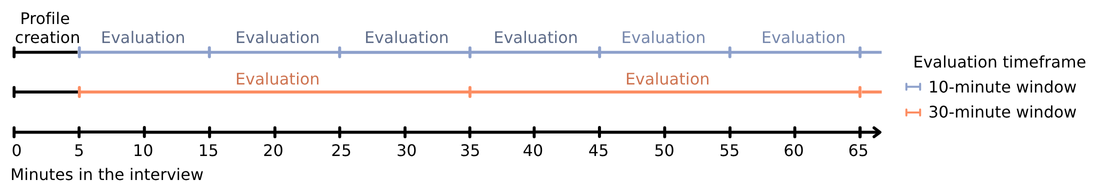}
\caption{Lexical profile evaluation strategy}
\label{fig:LP_eval}
\vfill
\end{figure}

To address the objectives of this study, lexical profiles were constructed once for each speaker using increasing segments from the beginning of the transcribed interview data. Specifically, profiles were created based on the first five, 10, 15, 20, 25, and 30 minutes of transcribed speech based on the annotated time indicators in the transcriptions. The varying timeframe lengths used to construct the profiles allowed us to explore how the amount of data used to construct the profiles affected their representativeness of the language use of the speaker, corresponding to the first aim of this study. To draw conclusions on the stability of the profiles over time, each of the constructed profiles was evaluated repeatedly across later timeframes of the interview data that was not used in the profile construction of the same speaker using 10- or 30-minute evaluation windows, corresponding to the second aim of the study. Figure \ref{fig:LP_eval} visualises this evaluation strategy for a profile created based on the first five minutes of speech. 

\subsection{Evaluation Metrics}
To evaluate the relevance and stability of the profiles over time, we measured exact and lemmatised repetition of profile vocabulary items and n-grams within the later evaluation timeframes, following common methods used to assess lexical alignment \cite{duplessis,duran2019align,menshikova2021lexical,sinclair2019tutorbot,koulouri}.

We computed recall and coverage defined as the proportion of vocabulary items or n-grams included in the lexical profile recurring in the later timeframe and the extent to which the lexical profile captured the vocabulary items or n-grams actually used in the later timeframe, respectively. They were calculated as follows:

\begin{equation}
    \mathrm{Recall} = \frac{|P_{\mathrm{items}} \cap E_{\mathrm{items}}|}{|P_{\mathrm{items}}|} \quad,\quad \mathrm{Coverage} = \frac{|P_{\mathrm{items}} \cap E_{\mathrm{items}}|}{|E_{\mathrm{items}}|}
\end{equation}

Here, \( P_{\mathrm{items}} \) and \( E_{\mathrm{items}} \) denote the set of vocabulary items (or n-grams) in the lexical profile and evaluation timeframe, respectively.

Following Duran et al. \cite{duran2019align}, we also included a cosine similarity measure to assess lexical alignment, incorporating word frequencies to capture the strength of repeated profile vocabulary. Cosine similarity was calculated between the frequency vectors \(\mathbf{P}\) and \(\mathbf{E}\), representing the frequency of vocabulary items in the union of the lexical profile and the later evaluation timeframe:

\begin{equation}
\mathrm{Cosine\ Similarity} = \frac{\sum_{i=1}^{n} P_i \cdot E_i}{\sqrt{\sum_{i=1}^{n} P_i^2} \cdot \sqrt{\sum_{i=1}^{n} E_i^2}}
\end{equation}

Note that cosine similarity was not calculated for the repetition of n-grams due to their sparse nature, especially in shorter texts. 

\section{Results}
\subsection{Profile Stability}
\begin{figure}[ht]
\centering
\includegraphics[width=\linewidth]{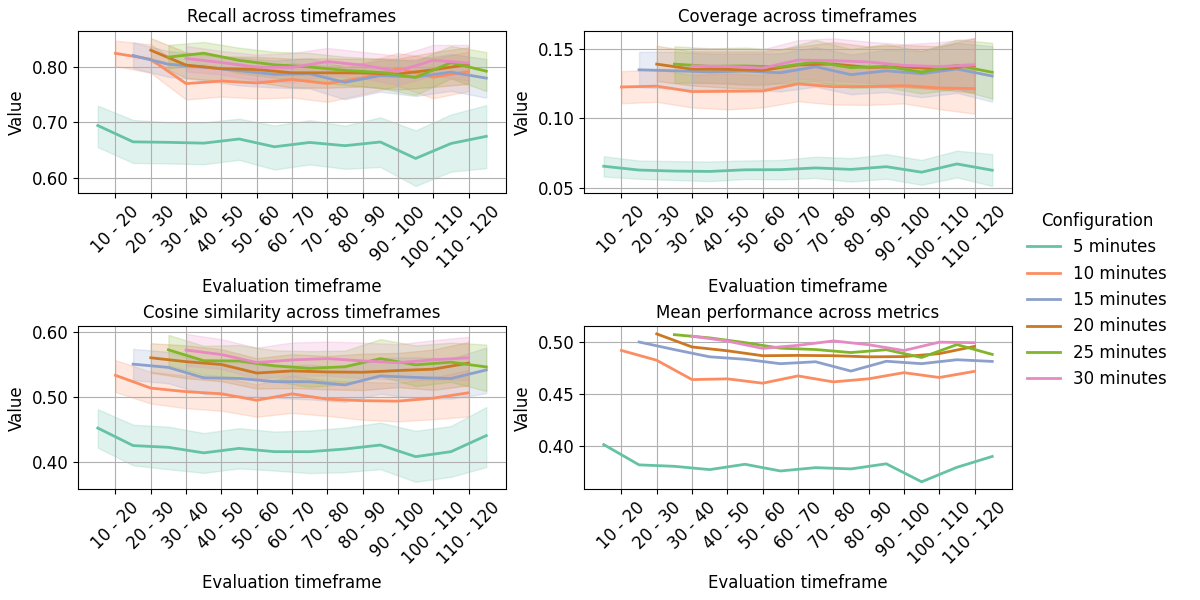}
\caption{Stability of lexical profiles across 10-minute evaluation timeframes}
\label{fig:Stability}
\vfill
\end{figure}

As shown in Figure \ref{fig:Stability}, evaluation metrics remained stable over time when evaluated repeatedly across the later 10-minute evaluation windows. Timeframes beyond 115 minutes were not included in the analysis due to data sparsity. Evaluation using the 30-minute windows is not shown, as it demonstrated similar stability with only a slight increase in recall and cosine similarity and a minor decrease in coverage. This is likely due to the increased total number of items in the evaluation timeframe, capturing more profile vocabulary items and n-grams while reducing the relative repetition frequency. Additionally, a clear underperformance was observed in the profiles created at the 5-minute timepoint, indicating that an early and small speech sample did not capture the representative lexical patterns of a speaker over time. Coverage was shown to be relatively low compared to the other metrics and the average performance across metrics, likely due to the limited number of vocabulary items per POS category and n-grams included in the profile compared to the full set of unique vocabulary items and n-grams in the evaluation timeframe, reducing the relative repetition frequency and therefore the coverage. However, the higher recall and cosine similarity indicated that the included vocabulary items and n-grams were indeed reused in the later evaluation timeframe.

\subsection{Part-of-Speech Configurations}
\begin{figure}[ht]
\centering
\includegraphics[width=\linewidth]{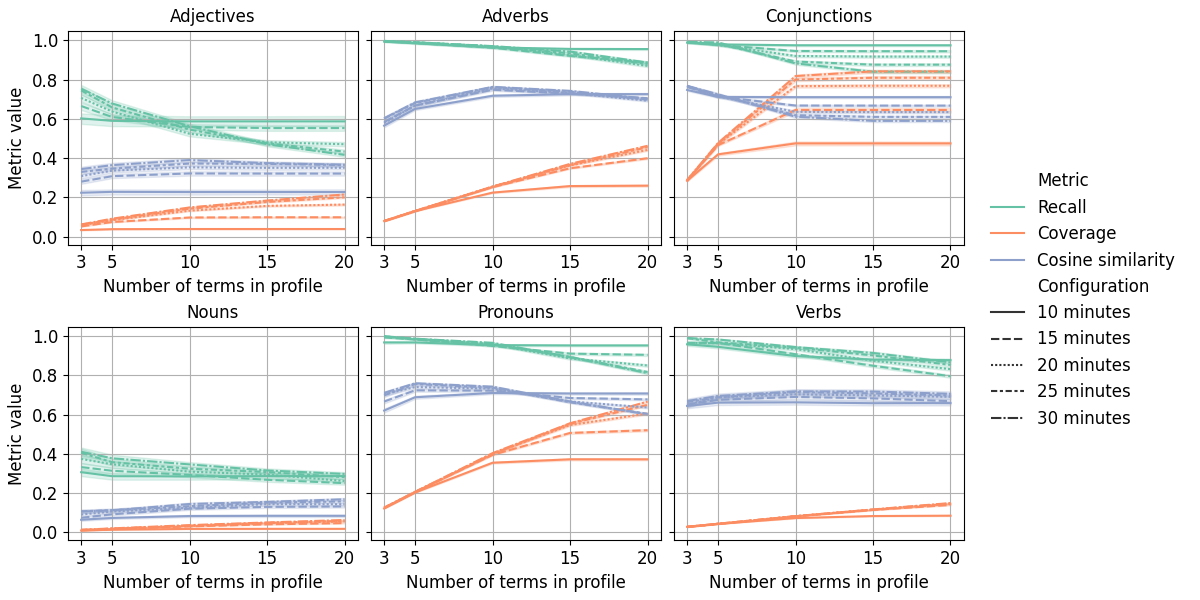}
\caption{Stability of lexical profiles across 10-minute evaluation timeframes per POS category}
\label{fig:Stability per POS}
\vfill
\end{figure}

The performance across different POS categories and profile configurations is shown in Figure \ref{fig:Stability per POS}. An increased number of vocabulary items per POS category included in the profiles generally increased the coverage, suggesting that larger profiles captured a broader range of the lexical patterns used by a speaker across time. However, recall tended to decrease for larger profiles, likely due to the inclusion of less stable and representative terms. Most POS categories reached a performance plateau for recall and cosine similarity with the inclusion of 5 to 10 terms, particularly for the profiles created at the 10-minute timepoint. Adjectives and conjunctions plateaued around 5 terms, whereas adverbs, nouns, pronouns, and verbs plateaued around 10 terms. These results suggested that profiles created at the 10-minute timepoint, including 5 items for adjectives and conjunctions and 10 items for adverbs, nouns, pronouns, and verbs each, offered a practical balance between the amount of data required and overall performance. In addition, adjectives and nouns showed a slight overall underperformance compared to the other categories, likely due to their high variability and sensitivity to the conversational topic, affecting their performance over time.

\subsection{Generalisation to Unseen Data}
\begin{figure}[ht]
\centering
\includegraphics[width=\linewidth]{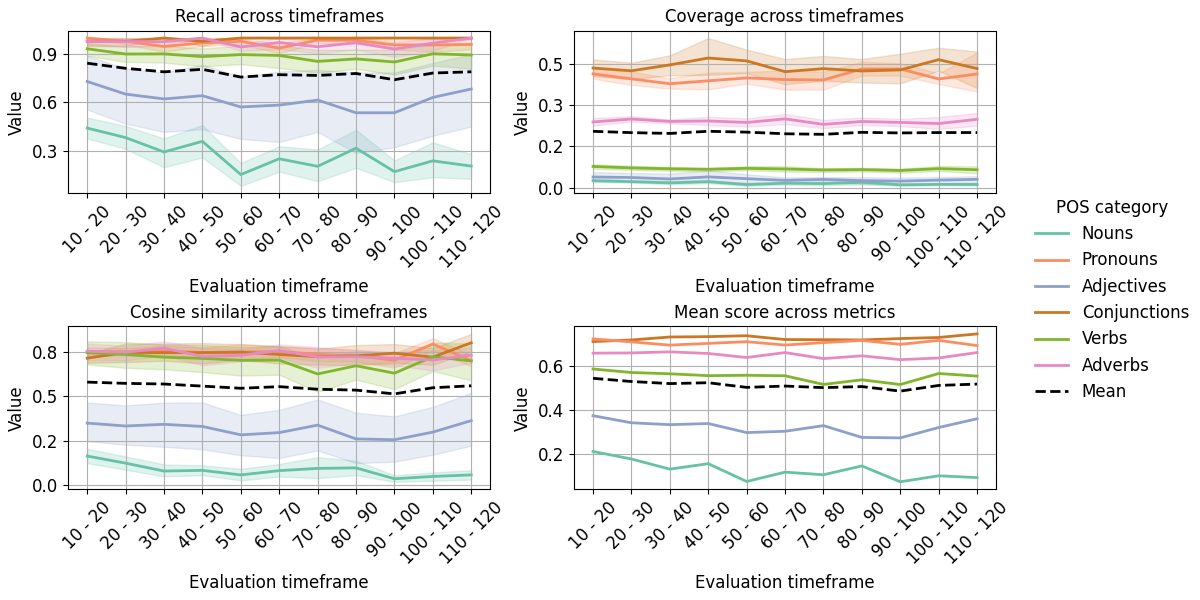}
\caption{Evaluation of the optimal profile configuration across evaluation timeframes}
\label{fig:Evaluation_timeframes}
\vfill
\end{figure}

The profile configuration that offered the best balance between performance and data requirements (constructed after 10 minutes of speech, including 5 items for adjectives and conjunctions and 10 items for adverbs, nouns, pronouns, and verbs each) was evaluated on unseen data. Specifically, profiles were constructed for the 10 interview transcripts in the holdout test set using this configuration and were repeatedly evaluated across the 10-minute evaluation windows. The performance across POS categories is shown in Figure \ref{fig:Evaluation_timeframes}. 

Overall, the results showed that this configuration generalised reasonably well to unseen data. The average performance across metrics closely resembled that of the training data (Figure \ref{fig:Stability}) and outperformed the 5-minute profiles constructed during training. Moreover, the performance metrics remained stable across evaluation timeframes, consistent with previous observations. 

Performance per POS category revealed decreased performance for nouns and adjectives, similar to the training data. Additionally, coverage for verbs was lower, likely due to the inclusion of the non-lemmatised verb forms in the lexical profiles, resulting in the profiles containing duplicate verb types. Increased variability in verb use in the evaluation timeframes may have led to the reduction in coverage scores. As in the training data, overall coverage remained relatively low, reflecting the limited vocabulary items and n-grams captured by the profiles compared to the full set of unique vocabulary items and n-grams present in the evaluation timeframe. However, higher recall and cosine similarity indicated the reuse of the included vocabulary items and n-grams in the profiles over time, supporting their relevance. 

\section{Discussion}
This study addressed the question: \textit{How can we construct stable, personalised lexical profiles that form the basis for lexical alignment in human-agent dialogue?} To investigate this, we systematically evaluated the stability and relevance of lexical profiles built from varying amounts of transcribed speech and different numbers of terms per POS category over time. 

Findings showed that profiles required at least 10 minutes of transcribed speech to achieve stable performance across recall, coverage, and cosine similarity. Recall and cosine similarity, reflecting the reuse and relevance of profile vocabulary items and n-grams, plateaued when profiles included 5 items for adjectives and conjunctions and 10 items for adverbs, nouns, pronouns, and verbs each. While increasing profile sizes improved coverage, it often resulted in decreased recall, likely due to the inclusion of less stable and representative items. Therefore, profiles constructed from 10 minutes of transcribed speech, containing 5 items for adjectives and conjunctions and 10 items for adverbs, nouns, pronouns, and verbs each, are considered to be a practical balance between performance and data requirements. The relatively low coverage for this configuration can be explained by the limited number of included vocabulary items and n-grams in the profile compared to the full vocabulary used in the evaluation timeframes. However, the consistent reuse of included vocabulary items and n-grams in the profiles over time, evidenced by stable recall and cosine similarity, implies the profiles' representativeness and temporal stability.

These results highlight the potential of such profiles to form the basis of alignment strategies in conversational agents. This becomes particularly relevant in contexts where extensive real-time user input cannot be guaranteed, such as interactions with individuals affected by dementia who often experience communication difficulties \cite{alzheimers-society-2024,banovic2018communication}. Due to the lack of a Dutch dementia speech dataset at the time of this study, future work should validate these findings using such data and also evaluate the generalisability across other speaker groups. In contrast to prompting techniques that heavily rely on continuous real-time user input \cite{srivastava2024}, these profiles move away from this dependency, enabling more inclusive lexical alignment strategies for speakers with communication difficulties. While the profiles showed stability and relevance over time, future research could explore selective updates based on key moments, such as changes in the speaker's cognitive state, which are known to potentially influence the lexical preferences of the speaker \cite{bucks-2000,hernandezdominguez-2018,rodriguez-2024,williams-2023}. These updates could help maintain the profiles' long-term relevance and stability while still providing a solid foundation for lexical alignment when little real-time user input is available. 

In summary, this study identified a lexical profile configuration that balances performance and data requirements, providing a solid foundation for lexical alignment strategies in conversational agents. Particularly in contexts where user input may be limited, inconsistent, or cognitively demanding, such as in dementia care, this approach allows for more inclusive lexical alignment strategies.

\section{Acknowledgments}
This work is part of the project Responsible AI for Voice Diagnostics (RAIVD) with file number NGF.1607.22.013 of the research program NGF AiNed Fellowship Grants, which is financed by the Dutch Research Council (NWO).

% Bibliography
\bibliographystyle{splncs04}
\bibliography{paper}

\end{document}